%% file: emnlp2022.tex
\pdfoutput=1

\documentclass[11pt]{article}

\usepackage[]{EMNLP2022}

\usepackage[hang,flushmargin]{footmisc}

\usepackage{times}
\usepackage{latexsym}
\usepackage{enumitem}
\usepackage{float}

\usepackage[T1]{fontenc}

\usepackage[utf8]{inputenc}

\usepackage{microtype}
\usepackage{booktabs}
\usepackage{soul}
\usepackage{multirow}

\usepackage{inconsolata}
\usepackage{graphicx}
\usepackage{tabularx}
\usepackage{amsfonts}
\usepackage{framed}
\usepackage{amsmath}
\usepackage{setspace}
\usepackage{newtxmath}

\newcommand{\method}{Resolved GPT-3}

\title{Large Language Models are Few-Shot Clinical Information Extractors}

\author{
Monica Agrawal\\
MIT CSAIL\\
\texttt{\small magrawal@mit.edu}
\And
Stefan Hegselmann\\
University of M{\"u}nster\\
\texttt{\small stefan.hegselmann@uni-muenster.de}\And
Hunter Lang\\
MIT CSAIL\\
\texttt{\small hjl@mit.edu}\AND
Yoon Kim\\
MIT CSAIL\\
\texttt{\small yoonkim@mit.edu}\\\And
David Sontag\\
MIT CSAIL\\
\texttt{\small dsontag@mit.edu}\\
}

\begin{document}
\maketitle
\begin{abstract}
\input{0_abstract}

\end{abstract}

\section{Introduction}
\input{1_introduction}

\section{Related Work}
\input{2_related}

\section{Methods}  \label{sec:methods}
\input{3_methods}

\section{Clinical Sense Disambiguation}
\input{4_task1disambiguation.tex}

\section{Biomedical Evidence Extraction}
\input{5_task2evidence.tex}

\section{Coreference Resolution} 
\input{6_task3coreference}

\section{Medication Extraction}
\input{7_task4medication}

\section{Conclusion}
\input{8_conclusion}

\input{9_limitations_ethics}
\bibliography{anthology,custom}
\bibliographystyle{acl_natbib}

\appendix

\clearpage
\input{10_appendix_examples}

\input{11_appendix_annotation}

\input{12_annotation_exp_details}

\input{13_appendix_cost}

\end{document}

%% file: 0_abstract.tex
A long-running goal of the clinical NLP community is the extraction of important variables trapped in clinical notes. 
However, roadblocks have included dataset shift from the general domain and a lack of public clinical corpora and annotations. 
In this work, we show that \textit{large language models}, such as InstructGPT \cite{ouyang2022training}, perform well at zero- and few-shot information extraction from clinical text despite not being trained specifically for the clinical domain. 
Whereas text classification and generation performance have already been studied extensively in such models, here we additionally demonstrate how to leverage them to tackle a diverse set of NLP tasks which require more structured outputs, including span identification, token-level sequence classification, and relation extraction. 
Further, due to the dearth of available data to evaluate these systems, we introduce new datasets for benchmarking few-shot clinical information extraction based on a manual re-annotation of the CASI dataset \cite{moon2014sense} for new tasks\footnote{\href{https://huggingface.co/datasets/mitclinicalml/clinical-ie}{https://huggingface.co/datasets/mitclinicalml/clinical-ie}}. 
On the clinical extraction tasks we studied, the GPT-3 systems significantly outperform existing zero- and few-shot baselines.

%% file: 1_introduction.tex
Clinical text contains a large amount of valuable information that is not captured by the structured data fields in electronic health records \cite{zweigenbaum2007frontiers, wang2018clinical}.
However, there are significant challenges to \emph{clinical information extraction}.
Because clinical text contains irregularities such as ambiguous jargon and nonstandard phrasal structure, most off-the-shelf NLP tools perform poorly, and clinical text annotation requires domain expertise \cite{zheng2011coreference}.
Further, due to the sensitive nature of clinical text, public corpora are rare and restrictively licensed. As a result, clinical NLP datasets tend to be small and splintered across institutions \cite{xia2012clinical}. \looseness=-1
To overcome these issues, practitioners often incorporate task-specific domain knowledge and regular expressions, even in modern deep learning pipelines, but these solutions can be brittle \cite{luo20202019, skreta2021automatically, chapman2001simple, irvin2019chexpert, johnson2019mimic, chauhan2020joint}. 
Modern systems that do not use some combination of these elements are generally limited to areas where labels are generated as a byproduct of normal clinical practice, such as ICD code prediction \cite{zhang2020bert} and mortality prediction \cite{si2019deep}.


\begin{figure}
\begin{center}
\includegraphics[scale=0.42]{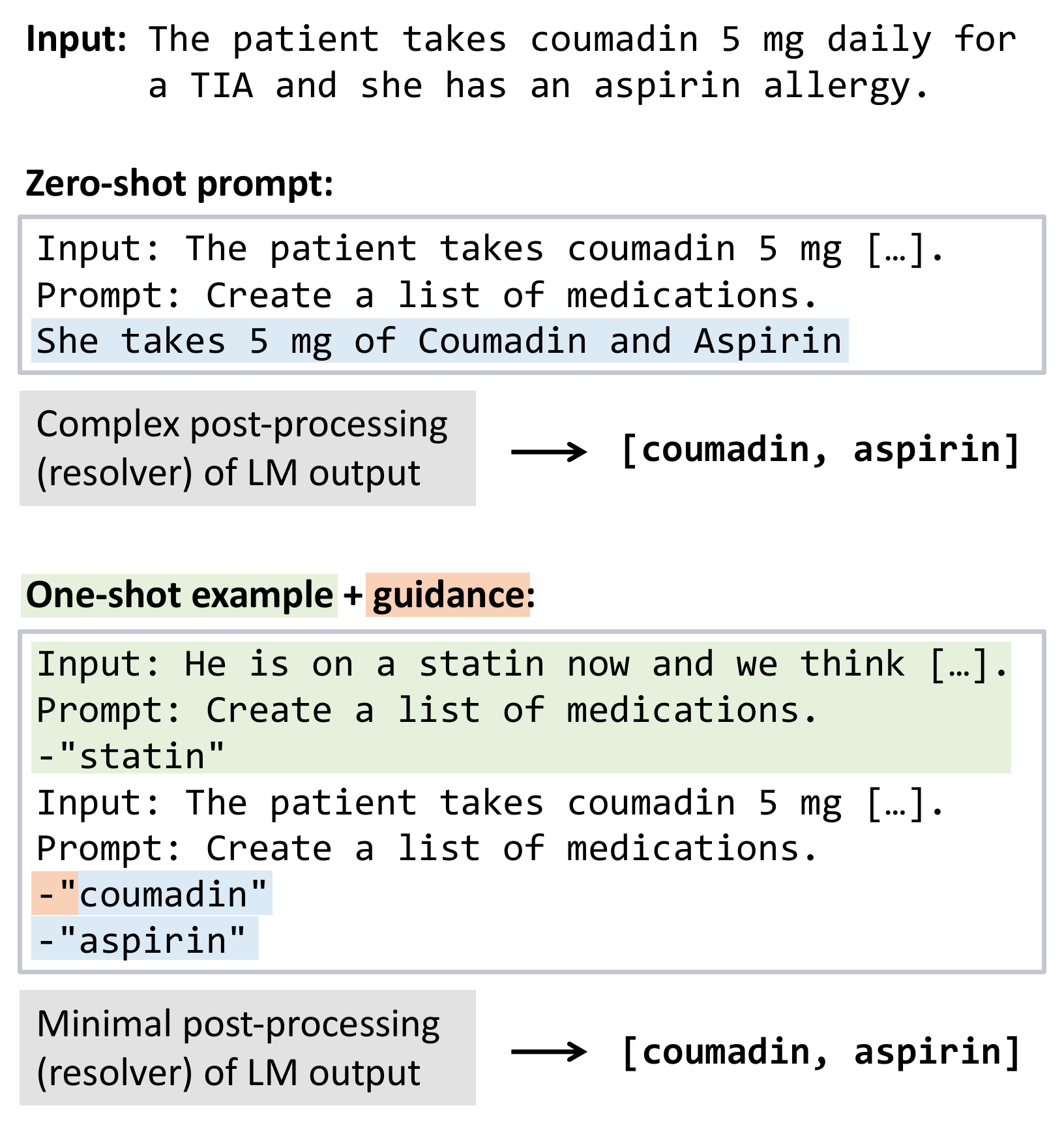}
\end{center}
\vspace{-0.8em}
\caption{Illustration of our approach using a one-shot example (green) and guidance (brown) to create a more structured LM output (blue). This significantly reduces the necessary post-processing effort of a resolver (gray).}
\label{fig:overview}
\vspace{-1em}
\end{figure}

\input{1b_overview_table}

\looseness=-1 In this work, we benchmark how large language models (LLMs) such as GPT-3 \cite{brown2020language, ouyang2022training} perform at clinical NLP tasks. This takes the form of three contributions:
\begin{itemize}[leftmargin=5pt,noitemsep]
    \item  We introduce \emph{three new annotated datasets} for benchmarking few-shot clinical information extraction methods, as many shared clinical corpora \cite{murphy2010serving, henry20202018, johnson2016mimic} have data use agreements that prevent their use with LLM APIs such as OpenAI's.
    The datasets were generated by re-annotating the dataset from \citet{moon2014sense} for new tasks.
    \item We show that GPT-3 performs well in clinical NLP over a set of diverse tasks (see Table \ref{tab:tasks_overview}), despite not being trained specifically for the domain.
    By replacing the complex hand-curated domain knowledge with the natural-language output of an LLM, the engineering effort required to solve a particular extraction task can be greatly reduced. 
    \item  While LLMs have been primarily evaluated at classification and generation tasks, our tasks involve a greater variety of expected output structures, such as relation extraction (see last three rows of Table \ref{tab:tasks_overview}). 
    We therefore introduce  \textit{guided prompt design} to steer the LLM towards an easy-to-structure output and  \textit{resolvers} to map from the LLM outputs to the structured label space; see Figure \ref{fig:overview}.
\end{itemize}



%% file: 1b_overview_table.tex
\newcolumntype{P}[1]{>{\raggedright\arraybackslash}p{#1}}
\begin{table*}[ht!]
\centering
\setlength{\tabcolsep}{6pt} 
\renewcommand{\arraystretch}{1.2} 
\small
\begin{tabular}{|P{1.8cm}|P{2.7cm}|P{2.7cm}|P{2.2cm}|P{3.8cm}|}
\hline
\textbf{Task} &
  \textbf{Description} &
  \textbf{Example Text} & \textbf{Answer} &
  \textbf{Data} \\ \hline\hline
Clinical sense \mbox{disambiguation} &
  Given a note and an abbreviation, expand the abbreviation \textcolor{violet}{(classification)} &
  \textit{[...] was sent to IR for thrombolysis. Post IR, ultrasound showed that [...]} & Interventional radiology &
  41 acronyms from 18,164 notes from CASI \cite{moon2014sense} and 8912 notes from MIMIC \cite{AdamsKBPE20} \\ \hline
Biomedical evidence extraction &
  Given an abstract, list interventions \textcolor{violet}{(multi-span identification/generation)} &
  \textit{[...] paliperidone extended- release tablets and  [...] with risperidone [...]} &-paliperidone extended-release tablets  -risperidone &
  187 abstracts (token-level) and 20 newly annotated abstracts (arm identification) from EBM-NLP \cite{nye2018corpus} \\ \hline
Coreference resolution &
  Given a note and a pronoun, identify the antecedent \textcolor{violet}{(span identification)} &
  \textit{[...] Did develop some tremors, however. These were well managed [...]} & some tremors &
  105 newly annotated examples from CASI \cite{moon2014sense} with one pronoun-antecedent pair each  \\ \hline
Medication status extraction &
  Given a note, extract medications and their status, e.g. active \textcolor{violet}{(NER + classification)} &
 \textit{[...] have recommended Citrucel  [...] discontinue the Colace. [...]} & -Citrucel:  \textit{active} \hspace{5pt} -Colace: \textit{discontinued} &
 105 newly annotated examples from CASI \cite{moon2014sense} with 340 medication-status pairs \\ \hline
Medication attribute extraction &
  Given a note, extract medications and 5 attributes, e.g. dosage, reason \textcolor{violet}{(NER + relation extraction)} &
 \textit{[...] she was taking 325 mg of aspirin per day for three years for a TIA. [...]}
 & aspirin: \{dose: 325 mg, freq: per day, duration: three years, reason: TIA\}&
 105 newly annotated examples from CASI \cite{moon2014sense} with 313 medications and 533 attributes \\ \hline
\end{tabular}
\caption{Overview of the five tasks studied in this work and the datasets that were used.}
\label{tab:tasks_overview}
\vspace{-0.8em}
\end{table*}

%% file: 2_related.tex
\subsection{Prompt-Based Learning}
\looseness=-1
In prompt-based learning (also known as in-context learning), a pretrained language model is adapted to different tasks via priming on natural language prompts---pieces of text that are combined with an input and then fed to the language model to produce an output for that task. 

This paradigm has been successful for few-shot and zero-shot learning at many general-domain tasks \cite{brown2020language,liu2021prompt,wei2021fine,sanh2021multitask}. 
More recently, large language models such as T0 and InstructGPT have re-configured their training objectives to explicitly encourage the model to perform well at such prompts \cite{sanh2021multitask, ouyang2022training}. 

While prompt-based learning can be extended straightforwardly to classification tasks (e.g., multiple choice), more complex tasks require creativity in their implementation \cite{mishra2021reframing}. 
For example, coreference resolution is often re-framed as classification, asking which of two antecedents a pronoun refers to \cite{sanh2021multitask} or whether a candidate antecedent is correct \cite{yang2022gpt}. 
This approach requires a list of antecedent candidates, which requires an additional component (e.g. a noun phrase generator) or many---potentially expensive---queries. 
Span classification and named entity recognition have been similarly reframed. 
For example, given a candidate entity \textit{X} and full model access, the entity type can be predicted via an argmax over the possible types \textit{Y} of the probability of statements like ``\textit{X} is a \textit{Y} entity'' \cite{cui2021template}. Alternatively, if only a single entity is being queried for a given input, prompting can be as simple as ``What is the location''\cite{liu2022qaner}; however, clinical NLP often concerns itself with extraction of multiple concepts. 
To extract multiple spans simultaneously, \citet{li2019entity} and \citet{li2019unified} use techniques from machine reading comprehension, relying on access to the underlying model and labeled data for training the extraction layer.
While InstructGPT \cite{ouyang2022training} has $\sim 2\%$ or $\leq 1k$ extraction examples in its training, the LLM output is never converted to a structured form, and extraction examples are only evaluated qualitatively for improvement over other models.
That is, only results for classification and generation tasks are quantified.

\subsection{Pretrained LMs for Clinical NLP}
Clinical text differs significantly from text typically utilized in general NLP, both in syntax and vocabulary \cite{wu2020deep}. As a result, the clinical NLP subcommunity often trains domain-specific models on clinical corpora following advances in language modeling from the broader NLP community. For example, clinical neural word embeddings were trained following word2vec \cite{mikolov2013distributed, wu2015clinical, roberts2016assessing}. More recently, following BERT, many clinical and biomedical variations swiftly followed including ClinicalBERT, SciBERT, BioBERT, and PubMedBERT  \cite{devlin2018bert, alsentzer2019publicly,ammar2018construction, lee2020biobert, gu2021domain}. 
However, in several applications, researchers observed the performance gains to be marginal to none over classical methods such as logistic regression \cite{chen2020intimate, krishna2021extracting}.
Additionally, previous work has so far been unable to achieve competitive results on {\em biomedical} NLP tasks using domain-agnostic LLMs like GPT-3 \cite{moradi2021gpt, gutierrez2022thinking}.

%% file: 3_methods.tex
\subsection{Predicting Structured Outputs with LLMs}

In this work, we assume only query access to a large language model (i.e., no gradients, no log probabilities). 

Suppose we have a set of $n$ examples $(\{x_i, a_i\})_{i=1}^n$, where $x_i$ is the input text as a string,  $a_i$ is (optional) side information as a string (e.g., which acronym to disambiguate). The outputs $y_i \in \mathbb{O}$ are unobserved (i.e., to be predicted).
The output space $\mathbb{O}$ is defined per task.
For example, for a binary sequence labeling task, if we let $\vert x_i \vert$ be the number of tokens in $x_i$, $\mathbb{O}$ is $\{0, 1\}^{\vert x_i \vert}$.

Prompt-based learning requires the specification of a prompt template to be applied on the input. In this work, we handcraft our prompt templates using a set of 5 validation examples per task. Let $p_j(x, a)$ be the result of filling prompt template $j$ with inputs $x$ and $a$, and further let $f(p_j(x, a)) \in \Sigma^{\star}$ be the string output by an LLM on input $p_j(x,a)$. 
The next step involves mapping the  LLM generation from $\Sigma^{\star}$ to the structured label space $\mathbb{O}$. 
For example, in classification, the \emph{verbalizer} defines a mapping between the LLM output space $\Sigma^\star$ and the discrete set of labels $\mathbb{O} = \{1,\ldots, L\}$ using a dictionary of token/label pairs \cite{schick2021exploiting}.
However, for our structured tasks of interest, the label space $\mathbb{O}$ is more complex, and more complicated functions are needed to map to an element of $\mathbb{O}$.
We define the \textit{resolver} $R$ as a function $R(x, a, f(p_1(x, a)))$ that maps the combined input and LLM output to the task-specific output space $\mathbb{O}$. For example, suppose the output space $\mathbb{O}$ is a \emph{list} of strings.
Then the resolver needs to turn each output $f(p_j(x, a))$ into a list (perhaps by choosing spans of text from inside of $f(p_j(x, a))$).
For example, for medication extraction we might have:

\vspace{-1.1em}
{\small
\begin{align*}
    &x = \textit{``switched Advil for Tylenol''},  a = \textit{``N/A''},\\
    &p_1(x,a) = \textit{``Note: switched Advil for Tylenol.''} \\
    & \hspace{47pt} \textit{Task: List medications.''} \\ 
    &f(p_1(x, a))= \textit{``Tylenol and Advil''} \\
    & R(x, a, f(p_1(x, a))) = \texttt{[``Tylenol'',``Advil'']}
\end{align*}}
\vspace{-1.2em}

We refer to the output from the resolver as \method{}, or \textbf{GPT-3 + R}, for short. Throughout, when comparing resolvers, we place in parentheses the lines of code (LOC) in the resolver, as a proxy for complexity (defined as human effort, not runtime). 
The required complexity of the resolver depends largely on the cleanliness of the prompt output, and by extension the prompt itself. 
We introduce \textit{guided prompt design} to simplify the resolver required for complex output. As seen in Figure \ref{fig:overview}, this consists of (i) a one-shot example with an output in the desired structured format (which could be incorrect content-wise), and (ii) guiding the model to use the same format. Specific constructions are found in Sections 6 and 7. 

\subsection{Dataset Annotation}
\label{subsec:annotation}
In the short-term, research on clinical extraction via prompting may rely on sending data to external APIs. 
Since data use agreements on many existing annotated clinical datasets prohibit such activity, there is a dearth of benchmarks for the community to build on. 
The de-identified Clinical Acronym Sense Inventory (CASI) dataset is therefore a valuable resource, as it is ``publicly available to support the research of the greater NLP and biomedical and health informatics community'' \cite{moon2014sense}. CASI contains snippets of clinical notes across specialties in four University of Minnesota-affiliated hospitals. While CASI was originally annotated for acronym disambiguation, we created three new annotated datasets from existing snippets of the CASI dataset. Annotation was performed by two of the authors who have background in both clinical NLP and medicine. For each task, a set of examples was jointly annotated to establish an annotation schema, each annotator then independently labeled the same set of 105 examples using PRAnCER software \cite{levy2021assessing}, and the two sets were then merged via joint manual adjudication. 

In the following sections, we show how to build simple resolvers for five clinical NLP tasks.
We find that resolvers for guided prompts are much easier to write than resolvers for un-guided prompts.
The implicit structured imposed by the prompt guidance means that resolvers for a guided prompt can be less than 10 LOC.
On the tasks below, we find that GPT-3 + R matches or exceeds strong few-shot, zero-shot, and even supervised baselines.

%% file: 4_task1disambiguation.tex
\vspace{1mm}\noindent\textbf{Overview.}
Clinical notes are rife with overloaded jargon and  abbreviations. \texttt{Pt} can mean patient, prothrombin time, physical therapy, or posterior tibial \cite{weeber2001developing, shilo2018analysis}. 
This ambiguity impacts the utility of notes for patients, clinicians, and algorithms \cite{kuhn2007abbreviations,mowery2016normalizing}. 
In this section, we first evaluate clinical sense disambiguation on the CASI dataset directly and then transfer a model distilled via weak supervision to another dataset.

\vspace{1mm}\noindent\textbf{Dataset 1.}
\looseness=-1
The Clinical Acronym Sense Inventory dataset consists of 500 text examples for each of 75 acronyms \cite{moon2014sense}. 
Due to noise in the dataset (e.g. duplications), it is common to filter to a subset of the dataset; we follow the filtering from \citet{AdamsKBPE20}, leading to a subset of 18,164 examples and 41 acronyms for evaluation. Similar to other works, we treat the task as multiple-choice.

\vspace{1mm}\noindent\textbf{Dataset 2.}
We additionally use a reverse substitution dataset \cite{AdamsKBPE20} generated over the MIMIC-III Critical Care Database \cite{johnson2016mimic}. In \textit{reverse substitution}, labeled data is generated from unlabeled text by replacing expansions (e.g. \textit{physical therapy}) with their acronyms (\textit{PT}) and using the original expansion as the label. 
We evaluate on their 8912 test examples over the same 41 acronyms as the CASI subset. Since we cannot query \textit{GPT-3} on this dataset, we distill and transfer a model trained on the outputs from Dataset 1.

\vspace{1mm}\noindent\textbf{Prompting + Resolver.}
We used \textit{GPT-3 edit} (using engine \textit{text-davinci-edit-001}) with greedy decoding (temperature $= 0$). 
For each example, we provided the full clinical snippet and appended the single instruction \texttt{Expand the abbreviation:\{abbr\}}. Since we did \emph{not} provide the LLM with the answer choices, the form of the output string could still differ slightly from all the candidate answers (e.g. editing ``RA'' to ``right atria'' when ``right atrium'' was expected). In the resolver, we choose the answer choice with the highest contiguous character overlap with the LLM generated output.

\vspace{1mm}\noindent\textbf{Model Distillation via Weak Supervision.}
\looseness=-1
Direct deployment of large language models can be difficult due to model size and data privacy.  
To remedy these issues, we follow several recent works \citep{lang2022co, smith2022language, wang2021want} and show that we can instead view the LLM + resolver system as a \emph{labeler} rather than as a \emph{classifier}, and that this can even boost performance.
In particular, we use outputs of this system on CASI as weak supervision \citep[e.g.,][]{ratner2017snorkel} to train a smaller, task-specific model. Here we fine-tune PubMedBERT \cite{gu2021domain} and follow \citet{lang2022co}; details and hyperparameters are found in the appendix.

\begin{table*}[t!]
    \centering
    \resizebox{1\textwidth}{!}{
    \begin{tabular}{lcccc}
        \toprule
         \textbf{Algorithm} & \textbf{CASI Acc.} & \textbf{CASI Macro F1} & \textbf{MIMIC Accuracy} & \textbf{MIMIC Macro F1} \\
         \midrule
         Random & 0.31  & 0.23 & 0.32  & 0.28 \\
         Most Common & 0.79  & 0.28 &  0.51  & 0.23  \\
         BERT (from \citet{AdamsKBPE20})  &  0.42  & 0.23 & 0.40  & 0.33 \\
         ELMo (from \citet{AdamsKBPE20})  &  0.55  & 0.38 & 0.58  & 0.53  \\
         LMC (from \citet{AdamsKBPE20})  &  0.71  & 0.51 & 0.74  & \textbf{0.69} \\
        \midrule
         \textit{GPT-3 edit} + R: 0-shot  & 0.86 & 0.69 & * &  *\\
         \textit{GPT-3 edit} + R: 0-shot + distillation
         & \textbf{0.90} & \textbf{0.76} & \textbf{0.78}  & \textbf{0.69} \\
         \bottomrule
    \end{tabular}}
    \caption{\textbf{Clinical sense disambiguation.} Accuracy and macro F1 for zero-shot language modeling approaches on a subset of the Clinical Acronym Sense Inventory (CASI) data set \cite{moon2014sense} and the MIMIC Reverse substitution dataset \cite{AdamsKBPE20}.
    GPT-3 is not run on MIMIC due to the data use agreement.
    To evaluate on MIMIC we distill GPT-3 + R into a smaller model by treating the outputs as weak supervision and following \citet{lang2022training} ``+ distillation'', then evaluate the smaller model on MIMIC as well.}
    \label{tab:disamb_results}
    \vspace{-0.5em}
\end{table*}

\vspace{1mm}\noindent\textbf{Baselines.} We compare the performance of our approach to other zero-shot language modeling methods: (i) Latent Meaning Cells (LMC), a deep latent variable model from \citet{AdamsKBPE20} which is pre-trained on millions of notes from MIMIC, (ii) ELMo pre-trained on the same dataset \cite{peters-etal-2018-deep},  and (iii) Clinical BioBERT \cite{alsentzer2019publicly}. Numbers for these three baselines are taken from \citet{AdamsKBPE20}; for all three, they choose the answer choice with the most similar representation to the contextual representation of the acronym. 
We also show performance for random guessing and choosing the most common answer choice per acronym (since the expansions of many acronyms follow a long-tailed distribution).

\vspace{1mm}\noindent\textbf{Evaluation.}
Accuracy and macro F1 are calculated per acronym and averaged over all acronyms (see left of Table \ref{tab:disamb_results}).
On CASI, GPT-3 edit + R alone already clearly outperforms the LMC model on both metrics, and the addition of weak supervision with PubMedBERT further boosts this performance. 
On the MIMIC Reverse Substitution dataset, despite being transferred to a new domain, our weakly-supervised PubMedBERT model performs similarly to LMC \cite{AdamsKBPE20}, which was pre-trained specifically on the MIMIC distribution.
This indicates we can use GPT-3 edit + R to label a public dataset, distill its labels into a smaller task-specific model, and then transfer that model to a private dataset to obtain competitive performance.
Since the CASI dataset is publicly accessible, one possible caveat is that the dataset could have been in the language model's training data; to investigate further (see Section \ref{sec:casi-not-in-training-set}), we prompt the LLM on acronyms \emph{not in the original annotations}.

%% file: 5_task2evidence.tex
\vspace{-1mm}
\textbf{Task Overview.     }
Evidence-based medicine (EBM) involves synthesizing findings from across clinical research studies, but the current rapid clip of research makes it nearly impossible to keep up with all studies \cite{sackett1997evidence, bastian2010seventy}. 
Therefore, automated approaches for parsing clinical abstracts could aid the adoption of EBM \cite{verbeke2012statistical, nye2018corpus}. Here, we focus on extracting interventions and controls (which we will refer to just as Intervention), where the underlying goal is to identify the distinct arms of a clinical trial \cite{nye2018corpus}.
Token-level classification is often used as a proxy for this goal, but distilling identified spans into distinct interventions is non-trivial and often requires significant domain knowledge.
Prior work on arm identification has attempted to use coreference resolution \cite{ferracane2016leveraging} and to identify of pairs of spans with redundant information \cite{nye2018corpus}.

\vspace{1mm}
\noindent\textbf{Dataset.} We assess intervention identification from the angles of (i) the token classification proxy task and (ii) the underlying task of arm identification.
For (i), we use the token-level annotations provided in version 2 of the dataset from \citet{nye2018corpus} and evaluate on the 187 test abstracts provided. The average Cohen's $\kappa$ was only 0.59 on this set. 
For (ii), the two annotators from Section \ref{subsec:annotation} manually derived a list of the intervention-control arms for 20 abstracts in the test set, with perfect agreement.

\vspace{1mm}
\noindent\textbf{Prompting + Resolvers.}
We use a single prompt with InstructGPT (engine \textit{text-davinci-002}) and greedy decoding. The resolver for the token-labeling task removes noisy tokens (stop words) from the LLM output, maps remaining tokens in the output to the original input and labels those as 1, and merges fractured spans. The full process can be found in Appendix \ref{apx:biomedical-evidence-extraction}. 
For the arm identification task, resolving simply involved splitting the output string on new lines.

\vspace{1mm}
\noindent\textbf{Comparison.} We compare to supervised approaches that train on the 4800 labeled training examples from \citet{nye2018corpus}. 
PubMedBERT with an additional classification layer (LSTM or CRF) achieves close to state-of-the-art performance on the full task \cite{gu2021domain}. 
Since prior works report numbers combined over multiple classes, we re-run training on only the Intervention label using PubMedBERT-CRF.
We also include the best supervised baseline from \citet{nye2018corpus}, an LSTM-CRF over word and character-level embeddings.

\vspace{1mm}
\noindent\textbf{Token-level results (Proxy Task).} We first evaluate sequence labeling precision at the token-level (F1 in Table \ref{tab:pico_results}).
\method{} performs respectably compared to supervised deep baselines, but underperforms on these token-level metrics. 
Many error modes occur due to particularities of the schema, e.g. including extra details (like dosing schedule or route of administration) and only including an acronym or its expansion, but not both. A clarifying example can be found in Section \ref{apx:biomedical-evidence-extraction}.

\begin{table}[t!]
    \centering
    \resizebox{0.5\textwidth}{!}{
    \begin{tabular}{lcc}
        \toprule
         \textbf{Algorithm} & \textbf{\shortstack{Token-level \\ F1}}  & \textbf{\shortstack{Abstract-level \\ Accuracy}} \\
         \midrule
         PubMedBERT-CRF (sup) & \textbf{0.69} & 0.35 \\
         LSTM-CRF (sup)  & 0.65 & * \\
         \midrule
         GPT-3 + R: 0-shot & 0.61 & \textbf{0.85} \\
         \bottomrule
    \end{tabular}}
    \caption{\textbf{Biomedical Evidence Extraction}. Test F1 scores on the binary token-level sequence labeling problem for Intervention identification  \citep{nye2018corpus}, and abstract-level accuracy at arm identification. The supervised baselines were trained on 4,800  abstracts.}
    \label{tab:pico_results}
    \vspace{-1em}
\end{table}

\vspace{1mm}
\noindent\textbf{Arm Identification Results.} 
To measure arm identification accuracy, we evaluated whether the number of arms was accurate and manually checked whether the main differentiator of each intervention arm was captured, similar to \citet{ferracane2016leveraging}. For the PubMedBERT baseline, in order to distill the identified spans to a list of arms, we assume (i) oracle splitting of spans into arms (given a span describing multiple arms, we can correctly split the span) and (ii) near-oracle coreference resolution (given multiple spans describing the same arm, we can correctly merge). \method{} successfully identified the correct number and content of the arms in 17 of the 20 examples. The three examples it missed were also missed by PubMedBERT. Assuming oracle splitting and coreference (a nontrivial task), PubMedBERT would still have issues with 10 further examples. Details of the evaluation and error modes are in Section \ref{apx:biomedical-evidence-extraction}.

%% file: 6_task3coreference.tex
\textbf{Task Overview. } Coreference resolution involves grouping noun phrases that refer to the same underlying entity (e.g. a person, a medical concept), and it is considered particularly important for clinically accurate information retrieval and summarization \cite{zheng2011coreference}. For example, when surfacing past medical history, it is critical to correctly parse pronouns to understand whether the history describes the patient or a family member.

\vspace{1mm}\noindent\textbf{Dataset Description.} 
In clinical NLP, coreference resolution has been largely evaluated on the 2011 i2b2/VA challenge, which consists of thousands of coreference \textit{chains}  \cite{uzuner2012evaluating}. 
Due to i2b2's data use agreement, the two annotators annotated a new dataset using CASI snippets, with 5 coreference pairs for prompt design and 100 pairs for evaluation \cite{moon2014sense}. 
We prioritized difficult examples by focusing on pronoun coreference, where the input is a pronoun, the output its antecedent, and no tokens overlap between the two. More details are in Section \ref{apx:coreference-resolution-dataset-details}.

\vspace{1mm}\noindent\textbf{Prompting and Resolvers.}
\looseness=-1 We used the 5 examples for prompt design with InstructGPT (engine \textit{text-davinci-002}) and greedy decoding (temperature $=0$). We use a guided 1-shot prompt, where we provide an example input and begin a formatted response: ``\{pronoun\} refers to''. For 1-shot, we experiment with both correct (the true noun phrase) and incorrect answers (a random incorrect noun phrase preceding the pronoun) in the example input to tease apart the effect of the example answer versus the example formatting. 
To clarify that effect, we average over results from 5 different 1-shot examples.
We also compare to an \textit{unguided} zero-shot prompt, which simply appends ``What does \{pronoun\} ... refer to?" to the input.
The zero-shot resolver involves mapping tokens back to the input due to potential paraphrases; the one-shot resolver involves only the removal of a single quotation mark, making the guided prompt easier to resolve.
Section \ref{apx:coreference-prompts} contains more detail on the prompts.

\begin{table}[t!]
    \centering
    \resizebox{0.5\textwidth}{!}{
    \begin{tabular}{lcc}
        \toprule
          \textbf{Algorithm} & \textbf{\shortstack{Recall}} & \textbf{\shortstack{Precision}}\\
         \midrule
         \citet{toshniwal2020learning, toshniwal2021generalization} &  $0.73_{\phantom{.01}}$ & $0.60_{\phantom{.01}}$ \\
         \midrule
         GPT-3 + R (50 LOC): 0-shot  & $\mathbf{0.78}_{\phantom{.01}}$ & $0.58_{\phantom{.01}}$  \\ 
         GPT-3 + R (1 LOC): 1-shot (incorrect) & $0.76_{.02}$  & $\mathbf{0.78}_{.04}$  \\ 
         GPT-3 + R (1 LOC): 1-shot (correct) & $0.75_{.04}$ & $0.77_{.04}$ \\ 
         \bottomrule
    \end{tabular}}
    \caption{\textbf{Coreference Resolution}. Macro unigram recall and unigram precision of methods on our newly annotated task using CASI \cite{moon2014sense}. The end-to-end baseline was trained on three non-clinical coreference resolution datasets and transferred to this new setting. 1-shot results are averaged over 5 prompts.}
    \label{tab:coref_results}
    \vspace{-1em}
\end{table}

\vspace{1mm}\noindent\textbf{Comparison.} 
We compare to deep end-to-end coreference resolution, as it has been shown to perform well \cite{lee2017end}.
In particular, we compare to the \textit{longdoc} model from \cite{toshniwal2020learning}, which trained on multiple coreference datasets in order to generalize to new settings. 

\vspace{1mm}\noindent\textbf{Results.}

We evaluated via macro unigram recall (\%  of label's unigrams in the resolved output) and unigram precision (\% of unigrams in the resolved output in the label) (Table \ref{tab:coref_results}).
We tokenized using Stanza \citep{qi2020stanza} for these metrics.
While the \textit{longdoc} baseline trained on thousands of non-clinical coreference examples performed considerably well already, it is outperformed by \method{}. We found the 1-shot example mostly constrains the LLM output to quoting (rather than paraphrasing); without guidance, the LLM may output e.g., ``The antecedent is unclear." Further, the accuracy of the 1-shot example was irrelevant to the performance, an observation previously reported in the classification setting, now seen for span extraction \cite{min2022rethinking}.

%% file: 7_task4medication.tex
\begin{table}[t!]
\resizebox{0.5\textwidth}{!}{
    \begin{tabular}{lcc}

        \toprule
        \textbf{Algorithm} & \textbf{Recall}   & \textbf{Precision} \\
        \midrule
        ScispaCy \cite{neumann-etal-2019-scispacy}  & $0.73_{\phantom{.01}}$         &  $0.67_{\phantom{.01}}$ \\
        \midrule
        GPT-3 + R (32 LOC) (0-Shot)                                  & $0.87_{\phantom{.01}}$  &  $0.83_{\phantom{.01}}$ \\
        GPT-3 + R (8 LOC)\ \ \ (1-Shot)                                & $\mathbf{0.90}_{.01}$ &  $\mathbf{0.92}_{.01}$\\
         \bottomrule
    \end{tabular}}
    \caption{\textbf{Medication extraction.} Micro recall and precision for medication extraction on our self-annotated dataset.}
    \label{tab:casi_medications_extraction_results}
\end{table}

\begin{table}[t!]
\resizebox{0.5\textwidth}{!}{
    \begin{tabular}{lcc}

        \toprule
         \textbf{Algorithm} & \begin{tabular}{@{}c@{}}\textbf{Conditional}\\
         \textbf{Accuracy}\end{tabular}   & \begin{tabular}{@{}c@{}}\textbf{Conditional}\\
         \textbf{Macro F1}\end{tabular} \\
         \midrule
         T-Few (20-shot)  &   $0.86_{\phantom{.01}}$    &  $0.57_{\phantom{.01}}$ \\
         \midrule
         GPT-3 + R (32 LOC) (0-Shot)    &   $0.85_{\phantom{.01}}$    &  $0.69_{\phantom{.01}}$ \\
         GPT-3 + R (8 LOC) (1-shot)                                      &  $\mathbf{0.89}_{.01}$  &  $0.62_{.04}$\\

         \addlinespace[1mm]
         \begin{tabular}{@{}c@{}}GPT-3 + R (8 LOC)  (1-shot)\\ + added classes\end{tabular} & $0.88_{.02}$   & $\mathbf{0.71}_{.03}$  \\
         \addlinespace[1mm]
         \begin{tabular}{@{}c@{}}GPT-3 + R (8 LOC)  (1-shot)\\ with shuffled classes\end{tabular} & $0.88_{.01}$    & $0.66_{.03}$  \\
         \bottomrule
    \end{tabular}}
    \caption{\textbf{Medication status classification.} Conditional accuracy and macro F1-score for Identification of medication status \textit{active}, \textit{discontinued}, and \textit{neither}.}
    \label{tab:casi_medications_classification_results}
    \vspace{-1em}
\end{table}

The recognition of clinical concept mentions (problems, treatments, etc.), their modifiers  (e.g., negation), and relations (e.g., dosage) is a fundamental building block in clinical NLP \cite{jiang2011study}. Here we examine the extraction of medication concepts with two different schemas.

\subsection{Recognition + Status Classification}
Here we extract a list of medications and label each with a status modifier: active, discontinued, or neither (e.g. allergy or proposed medication).

\vspace{1mm}\noindent\textbf{Dataset description.}
We created new annotations for medication and status on top of CASI \citet{moon2014sense}. The examples were enriched for changeover in treatment.  
For 105 randomly selected snippets, the annotators extracted all medications mentioned and classified its status with one of the 3 labels. Further details are in Appendix \ref{apx:medication-status-extraction-dataset-details}. Unlike in Section \ref{sec:lens2}, all mentions corresponding to the same medication are collapsed.

\vspace{1mm}\noindent\textbf{Prompting and Resolver.} We again used 5 examples for prompt design with InstructGPT (engine \textit{text-davinci-002}) and greedy decoding.
Our prompt asked the model to simultaneously output the list of medications and the status of each. We evaluate the prompt in an unguided zero-shot manner and in a guided one-shot manner. Further, to clarify the effect of the 1-shot example on modifier accuracy, we examine how status classification performance changes if we (i) artificially augment the 1-shot example so all three status classes are observed, and (ii) whether the statuses need to be correct, or just present. We averaged over 5 different 1-shot inputs to clarify these effects; each 1-shot example contained between 3 and 8 medications. 
We describe the resolvers for the zero- and one-shot cases in detail in Section \ref{apx:medication-status-extraction-resolver}; the former involved several regular expressions, and the latter required only a few short lines.

\begin{table*}[t!]
\resizebox{1\textwidth}{!}{
    \centering
    \begin{tabular}{llccccccc}
        \toprule
        \textbf{Subtask} & \textbf{Algorithm} & \textbf{Medication} & \textbf{Dosage} & \textbf{Route} & \textbf{Frequency} & \textbf{Reason} & \textbf{Duration} \\
        \midrule
        \multirow{2}{*}{Token-level} & PubMedBERT + CRF (Sup.) & 0.82 & 0.92 & 0.77 & 0.76 & 0.35 & \textbf{0.57} \\
                   & GPT-3 + R: 1-shot    & \textbf{0.85} & 0.92 & \textbf{0.87} & \textbf{0.91} & \textbf{0.38} & 0.52 \\
                \midrule
        \multirow{2}{*}{Phrase-level} &  PubMedBERT + CRF (Sup.) & 0.73 & 0.78 & 0.71 & 0.41 & \textbf{0.22} & \textbf{0.30} \\
                   & GPT-3  + R: 1-shot      & \textbf{0.75} & \textbf{0.82} & \textbf{0.81} & \textbf{0.87} & 0.21 & 0.25  \\
        \midrule
        \multirow{2}{*}{Relation Extraction} & \begin{tabular}{@{}l@{}}PubMedBERT + CRF + \\ \citet{shi2019simple} (Sup.)\end{tabular} & * & 0.67 & \textbf{0.65} & 0.36 & 0.19 & \textbf{0.21}  \\
        \addlinespace[1.5mm]
                   & GPT-3  + R: 1-shot    & * & \textbf{0.80} & 0.63 & \textbf{0.60} & \textbf{0.34} & 0.16  \\        
         \bottomrule
    \end{tabular}}
    \caption{\textbf{Medication attribute extraction.} F1 scores on our newly annotated medication extraction dataset.
    The baselines are trained using supervised learning on i2b2 \citep{uzuner2010extracting}, then transferred to the test domain.
    \emph{Relation Extraction} additionally requires the model to match modifiers (dosage, route, etc.) to the medication span.    
    Baseline end-to-end relation extraction performance suffers due to errors cascading from the extraction step.}
    \label{tab:med_relation_results}
    \vspace{-1em}
\end{table*}

\vspace{1mm}\noindent\textbf{Comparison.}
We used a rule-based method as a medication extraction baseline, since historically they perform well \cite{sohn2014medxn}. To this end, we leveraged the Python library ScispaCy with the \texttt{en\_core\_sci\_sm} package for entity recognition \citep[][details in Appendix \ref{apx:medication-status-extraction-resolver}]{neumann-etal-2019-scispacy}.\footnote{We do not use a supervised baseline trained on the i2b2 2009 challenge data (as in Section \ref{sec:lens2}) because their schema purposefully excluded medications in the \textit{Neither} category.}
For medication status classification, we compare to \texttt{T-Few} \citep{liu2022few}, a few shot LLM method fine-tuned on a set of additional snippets we labeled from the same distribution (20 snippets containing 60 medication statuses). This method predicts the status, \emph{given the token indices for each medication}.

\vspace{1mm}\noindent\textbf{Results.\ }
Table \ref{tab:casi_medications_extraction_results} shows micro recall and precision for medication extraction; we count a prediction as correct if the predicted string exactly matches one.
Overall, \method{}  outperforms the ScispaCy linkage baseline consistently by a considerable margin. 
The addition of the 1-shot example greatly improves precision, since in the 0-shot case, some GPT-3 outputs included extraneous extractions (e.g. a procedure). 
Typical failure modes of the baseline include incorrect recognition of overloaded abbreviations and missing vendor-specific drug names. 
Table \ref{tab:casi_medications_classification_results} shows \textit{conditional} accuracy on medication status classification. For an apples-to-apples comparison, we conditioned on the subset of medications found by all GPT-3 methods (241/340) and evaluated T-few on that subset as well. We find that if the rarer \textit{Neither} class wasn't demonstrated in the 1-shot example, it was unlikely to be output, depressing the F1 score; including all classes in the 1-shot prompt appears more important than necessarily having the correct labels.

\subsection{Recognition + Relation Extraction}
\label{sec:lens2}
\noindent\textbf{Dataset description.}
The annotators created a second new dataset for medication extraction from the snippets from \citet{moon2014sense}. The annotators closely followed the schema from the 2009 i2b2 medication challenge \cite{uzuner2010extracting}, with small deviations explained in Appendix \ref{apx:med-attribute-annotation}. For 105 randomly selected snippets, the annotators labeled mentions of medications, dosages, routes, frequencies, reasons, and durations, if available, and their correspondences. We examine the task from three different framings: a token-level annotation task, a phrase-level annotation task, and end-to-end relation extraction. For the example phrase ``Tylenol twice daily'', the desired output for the tasks would be: \textit{[Med, Frequency, Frequency]}, \textit{[B-Med, B-Frequency, I-Frequency]}, and \textit{{Medication: "Tylenol", Frequency: "twice daily".}}, respectively.

\vspace{1mm}\noindent\textbf{Prompting and Resolver.} We again used 5 examples for prompt design with InstructGPT (engine \textit{text-davinci-002}) and greedy decoding (temperature $= 0$). We use a different guided 1-shot prompt (containing 7 entities each) for each of the three framings outlined above; these can be found in Appendix \ref{apx:prompts_and_sample}. The resolvers for all were short.

\vspace{1mm}\noindent\textbf{Comparison.}
For token and phrase-level classification, we used a PubMedBERT model topped with a CRF layer.
For end-to-end relation extraction, we first used the token-level baseline to extract entity spans, then used the technique from \citet{shi2019simple} to classify whether each pair of entities was related.
We then postprocessed these pairwise outputs to match modifiers to their medications.
For all the three tasks, since we followed the 2009 i2b2 medication extraction annotation guidelines, we fine-tuned the baselines with labeled data from i2b2 (10 fully annotated notes with 154 medication mentions, which we postprocess into smaller annotated chunks) and directly evaluated them on our datasets. \citep{uzuner2010extracting}. 
Appendix \ref{apx:medication-baselines} contains more detail for the baselines and evaluation.

\vspace{2mm}
\noindent\textbf{Results.\ }
Table \ref{tab:med_relation_results} shows that the 1-shot GPT-3+R outperforms the i2b2-supervised baseline across all task framings. 
The baseline end-to-end relation extraction performance suffers due to cascading extraction errors, as the longest token in the medication name had to be matched.
GPT-3+R struggles with the \textit{duration} and \textit{reason} entities; however, it has been previously found that there is often large disagreement (F1 estimated 0.2--0.5) in inter-annotator agreement for these two entities, since they tend to be longer with ambiguous boundaries.

%% file: 8_conclusion.tex
In this work, we introduced new annotated datasets to show that (i) large language models have great promise at diverse clinical extraction tasks and (ii) we can guide generations to map to complex output spaces with only light post-processing. We also demonstrated how weak supervision over the system's outputs can be used to train smaller, task-specific models that are more deployable. The scope of clinical NLP extends past what we studied here, and important next steps involve experimenting with LLMs such as OPT \cite{zhang2022opt} for which we can run inference locally, enabling evaluation on existing benchmarks and fine-tuning. Another important direction involves leveraging the outputs from several prompts (e.g. 1-shot prompts with different examples) to learn to determine when GPT-3 is uncertain; this increased reliability will be vital given the high-stakes in clinical information extraction. Taken as a whole, our work indicates a new paradigm for clinical information extraction---one that can scale to the lofty goals of clinical NLP.

%% file: 9_limitations_ethics.tex
\section*{Limitations}

While large language models show great promise at clinical information extraction, there are clear limitations to their use. 
First, it is still difficult to guide a LLM to match an exact schema---clinical annotation guidelines are often multiple pages.
We found that even when the \method{} outputs were impressive qualitatively, they did not always match at the token-level. For example, in tagging durations, one \method{} output was ``X weeks'' instead of ``for X weeks''. While this particular omission is trivial, it highlights the difficulty of communicating nuanced guidelines. 

Second, we found a bias in GPT-3 towards outputting a non-trivial answer even where none exists. For example, for medication extraction the prompt we ended up using was, ``Create a bulleted list of which medications are mentioned and whether they are active, discontinued, or neither.'' However, prior to this we had experimented with two separate prompts: ``Create a bulleted list of \textit{active} medications, if any.'' and ``Create a bulleted list of \textit{discontinued} medications, if any.'' 
If there was one active and one discontinued medication, the respective LLM outputs would be correct. 
However, if there were two active medications and none discontinued, the LLM primed with the discontinuation prompt tended to try to find an output and usually resorted to listing one or more active medications. 
Therefore, it is important to craft prompts or tasks that avoid this pitfall. 
For example, this could be achieved via (i) chaining multiple prompts, e.g., first asking if a certain entity type exists in the input, before asking for a list \cite{li2019entity, wu2022ai} or (ii) using an output structure like the sequence tagging approach. 

Finally, because of the data use restrictions on most existing clinical datasets, which prohibit publicly sharing the data (e.g., to the GPT-3 APIs), all tasks except for biomedical evidence extraction were derived from the publicly-available CASI dataset \cite{moon2014sense}. While we show the promise of transferring to a new setting in Section 4, it would be ideal to have been able to directly evaluate on multiple hospital systems at multiple points throughout time. Clinical text in CASI was drawn from notes from several hospitals and a diverse set of specialties, but is by no means representative of all clinical text. For example, the CASI paper states that the notes were ``primarily verbally dictated and transcribed,'' but this practice is not universal.  Further, as is unfortunately common in clinical NLP, we only tested in English, leaving testing the ability of LLMs to operate in different languages to future work \cite{neveol2018clinical}.

\section*{Ethics Statement}

The datasets introduced in this paper involved only new annotations on top of existing, publicly available clinical text. Dataset annotation was conducted by two authors of the paper, and therefore there are no associated concerns, e.g. regarding compensation. As discussed in limitations, we believe these new annotated datasets serve as a starting point for the evaluation of LLMs on clinical text, but we concede that conclusions about specific performance cannot be ported to other languages, hospital systems, or temporal settings (as clinical text is quite subject to dataset shift).

If large language models were to be integrated into clinical extraction pipelines, as presented in this paper, there are large potential benefits. Clinical text is being created at a scale far too large for manual annotation, and as a result, cohorts for clinical study are largely small and hand-curated. Automatic structuring of clinical variables would help catalyze research that may be prohibitively expensive otherwise -- allowing for study of rarer or less funded diseases as well as the analysis of real-world evidence for subpopulations that may not be observed in clinical trials. However, due to the high-stakes setting, it is imperative that the performance of such a system is evaluated in the same environment it will be used in, and that the performance numbers are stratified by cohorts of note (e.g. racial, socioeconomic, patient comorbidities, disease stage, site of care, author's clinical role and seniority); such variables were not available in the data we used here.

In this work, we accessed the GPT-3 model using the OpenAI API alone. However, we acknowledge that even the inference cost is still nontrivial (see Appendix \ref{apx:experimental_cost}). We presented in Section 4 a paradigm of using weak supervision to distill a much smaller model, using pseudolabels learned from GPT-3, and we encourage such work to mitigate the environmental impact of deployment.

\clearpage

\section*{Acknowledgements}
MA was supported by a Takeda Fellowship, the MIT Deshpande Center, and the MLA@CSAIL Initiatives which includes companies Arrow, EY, Cisco, Capital One, Ahold Delhaize, SAP, and BT. HL was supported by NSF AiTF award CCF-1723344, and SH by the
German Academic Exchange Service. MA and DS were also partially supported by Memorial Sloan Kettering Cancer Center. Thanks to NVIDIA Corporation for their donation of two
NVIDIA A100 GPUs, and to OpenAI for providing quota to access their models. Finally, thanks to Rebecca Boiarsky for feedback on drafts of this paper.

%% file: 10_appendix_examples.tex
\section{Prompts and Sample GPT-3 Outputs}
\label{apx:prompts_and_sample}

\definecolor{shadecolor}{gray}{.95}
\newcommand{\hll}[1]{\textcolor{brown}{#1}}
\newcommand{\hlll}[1]{\textcolor{green}{#1}}
\newcommand{\hllll}[1]{\textcolor{red}{#1}}
\newcommand{\hlbox}[1]{\begin{shaded}\texttt{#1}\end{shaded}}

\newcommand{\lmout}[1]{\sethlcolor{cyan}\hl{#1}}
\newcommand{\exbox}[1]{
{\setstretch{0.8}
\noindent
\fbox{
\begin{minipage}{0.482\textwidth}{
\begin{flushleft}
\small{
\texttt{\noindent #1
}}
\end{flushleft}}
\end{minipage}}
}}

We present examples for each task alongside their corresponding prompts to illustrate different prompting strategies used for each task.


\subsection{Clinical Sense Disambiguation}
\label{apx:clinical_sense_disambiguation}

For clinical sense disambiguation we used GPT-3 edit mode which edits the input text rather than continuing it.


\subsubsection*{Example 1}

GPT-3 edit input:

\exbox{GENERAL: Patient is sedated on vent. HEENT: Normocephalic, atraumatic. Pupils are sluggish but reactive and equal bilaterally. NECK: Supple. No lymphadenopathy. No JVD. CARDIAC: Regular rate and rhythm. No murmurs. LUNGS: \hll{CTA}, intubated. ABDOMEN: Obese, nontender, positive bowel sounds. EXTREMITIES: Positive pulses, positive edema. SKIN: Chronic changes pretibial area bilaterally.}

GPT-3 edit prompt:

\exbox{Expand the abbreviation: CTA}

GPT-3 edit output: 

\exbox{GENERAL: Patient is sedated on vent. HEENT: Normocephalic, atraumatic. Pupils are sluggish but reactive and equal bilaterally. NECK: Supple. No lymphadenopathy. No JVD. CARDIAC: Regular rate and rhythm. No murmurs. LUNGS: \hll{Clear to auscultation}, intubated. ABDOMEN: Obese, nontender, positive bowel sounds. EXTREMITIES: Positive pulses, positive edema. SKIN: Chronic changes pretibial area bilaterally.}

\subsubsection*{Example 2}

GPT-3 edit input:

\exbox{2. Severe hypertension. 3. Severe mitral regurgitation. 4. Osteoporosis. PROCEDURES: 1. Coronary angiography and hemodynamic evaluation with right heart catheterization. Right heart catheterization shows right atrial pressure of 1, right ventricular pressure of 47/11, \hll{PA} pressure of 48/16, with a pulmonary capillary wedge pressure of 29, with a large B-wave confirmed with a wedge saturation of 95}

GPT-3 edit prompt:

\exbox{Expand the abbreviation: PA}

GPT-3 edit output: 

\exbox{2. Severe hypertension. 3. Severe mitral regurgitation. 4. Osteoporosis. PROCEDURES: 1. Coronary angiography and hemodynamic evaluation with right heart catheterization. Right heart catheterization shows right atrial pressure of 1, right ventricular pressure of 47/11, \hll{pulmonary artery} pressure of 48/16, with a pulmonary capillary wedge pressure of 29, with a large B-wave confirmed with a wedge saturation of 95\%.}

\vspace{2cm}
\clearpage
\subsection{Biomedical Evidence Extraction}
\label{apx:biomedical_evidence_extraction}

For biomedical evidence extraction we used GPT-3 Davinci. The generated output is shown in blue.


\subsubsection*{Example 1}

GPT-3 Davinci:

\exbox{Assessment of acute bronchodilator effects from specific airway resistance changes in stable COPD patients.\newline
BACKGROUND In COPD patients, reversibility is currently evaluated from the changes of forced expiratory volume at 1s ($\Delta$FEV1) and forced vital capacity ($\Delta$FVC). By lowering peripheral airway smooth muscle tone, bronchodilators should decrease dynamic hyperinflation, gas trapping, and possibly dyspnea at rest. Hence, we hypothesize that specific airway resistance changes ($\Delta$sRAW) should better characterize the acute response to bronchodilators. \newline
METHODS On two days, 60 COPD patients underwent dyspnea evaluation (VAS score) and pulmonary function testing at baseline and one hour after placebo or 300$\mu$g indacaterol administration. \newline
RESULTS Spirographic and $\Delta$sRAW-based criteria identified as responders 24 and 45 patients, respectively. $\Delta$sRAW correlated with changes of intrathoracic gas volume ($\Delta$ITGV) (r=0.61; p<0.001), residual volume ($\Delta$RV) (r=0.60; p<0.001), $\Delta$FVC (r=0.44; p=0.001), and $\Delta$VAS (r=0.73; p<0.001), while $\Delta$FEV1 correlated only with $\Delta$FVC (r=0.34; p=0.008).
Significant differences in terms of $\Delta$ITGV (p=0.002), $\Delta$RV (p=0.023), and $\Delta$VAS (p<0.001) occurred only if patients were stratified according to $\Delta$sRAW. [...]
\newline \newline
Create a bulleted list of the arms in this trial.
\newline \newline
\lmout{- Placebo} \newline
\lmout{- Indacaterol (300 $\mu$g)}
}

\subsubsection*{Example 2}

GPT-3 Davinci:

\exbox{Serum biochemical characteristics of Beluga, Huso huso (L.), in response to blood sampling after clove powder solution exposure. \newline
In order to investigate the effect of anesthesia on serum parameters, Beluga, Huso huso (L.) were blood-sampled immediately without anesthesia (control) or subjected to following anesthesia procedure: 40, 120, and 240 s exposure to 3,000, 700, and 500 mg l$^{-1}$ clove solution, respectively. Blood samples were collected after these periods, when fish were immobile and reached stage 4 anesthesia. Results showed that cortisol and glucose levels were significantly high in 700 and 500 but not 3,000 mg l$^{-1}$ group compared to control. Serum lactate levels were significantly high in 500 mg l$^{-1}$ group compared to control group. Lactate levels were not significantly differed between control, 3,000, and 700 mg l$^{-1}$ groups.
There were no significant differences in serum levels of cholesterol, total protein, lactate dehydrogenase, aspartate aminotransferase, alanine aminotransferase, Na$^{+}$, Cl$^{-}$, K$^{+}$, and Ca$^{2+}$. Results suggest that rapid anesthesia with higher dose is better than slow anesthesia with lower dose for blood sampling in Beluga.
\newline \newline
Create a bulleted list of the arms in this trial.
\newline \newline
\lmout{- Control} \newline
\lmout{- 3,000 mg l$^{-1}$ clove solution} \newline
\lmout{- 700 mg l$^{-1}$ clove solution} \newline
\lmout{- 500 mg l$^{-1}$ clove solution}
}

\clearpage
\subsection{Coreference Resolution}
\label{apx:coreference_resolution}
\label{apx:coreference-prompts}

\subsubsection*{Example 1: Zero-shot unguided}

GPT-3 Davinci:

\exbox{
[...] Her current regimen for her MS is Rebif Monday, Wednesday, and Friday and 1 gram of methylprednisolone p.o. every month. This had been working previously; however, she feels that her symptoms return before her next dose of methylprednisolone is due.
\newline \newline
What does "This" in the sentence "This had been working previously; however, she feels that her symptoms return before her next dose of methylprednisolone is due." refer to?
\newline \newline
\lmout{The sentence is unclear, but it seems to be referring to the Rebif regimen.}
}

\subsubsection*{Example 1: One-shot guided (incorrect)}

\exbox{
[...] ASSESSMENT \& PLAN: The patient has no physical evidence of progression of disease. It is somewhat worrisome that her CEA is up, but will be keep watching that to see if it is just within a normal swing for her. [...]
\newline \newline
What does "that" in the sentence "It is somewhat worrisome that her CEA is up, but will be keep watching that to see if it is just within a normal swing for her." refer to?
\newline \newline
"that" refers to "progression of disease"
\newline \newline
[...] Her current regimen for her MS is Rebif Monday, Wednesday, and Friday and 1 gram of methylprednisolone p.o. every month. This had been working previously; however, she feels that her symptoms return before her next dose of methylprednisolone is due.
\newline \newline
What does "This" in the sentence "This had been working previously; however, she feels that her symptoms return before her next dose of methylprednisolone is due." refer to?
\newline \newline
"This" refers to "\lmout{her current regimen for her MS"}
}

\subsubsection*{Example 1: One-shot guided (correct)}

\exbox{
[...] ASSESSMENT \& PLAN: The patient has no physical evidence of progression of disease.
It is somewhat worrisome that her CEA is up, but will be keep watching that to see if it is just within a normal swing for her. [...]
\newline \newline
What does "that" in the sentence "It is somewhat worrisome that her CEA is up, but will be keep watching that to see if it is just within a normal swing for her." refer to?
\newline \newline
"that" refers to "her CEA"
\newline \newline
[...] Her current regimen for her MS is Rebif Monday, Wednesday, and Friday and 1 gram of methylprednisolone p.o. every month. This had been working previously; however, she feels that her symptoms return before her next dose of methylprednisolone is due.
\newline \newline
What does "This" in the sentence "This had been working previously; however, she feels that her symptoms return before her next dose of methylprednisolone is due." refer to?
\newline \newline
"This" refers to "\lmout{her current regimen for her MS"}
}

\clearpage

\subsection{Medication Status Extraction}
\label{apx:medication_status_extraction}

\subsubsection*{Example 1: Zero-shot guided}

\exbox{
[...] home dose of Kadian as this is her long-acting medication and DC the continuous Dilaudid given IV. 5. Urinary tract infection with Klebsiella and E. coli, both sensitive to Levaquin. Since this was diagnosed Foley has been DC'd. For now would continue Levaquin and recheck urinalysis.
\newline \newline
Create a bulleted list of which medications are mentioned and whether they are active, discontinued, or neither.
\newline \newline
-"\lmout{Kadian" (active)}
\newline
\lmout{-"Dilaudid" (discontinued)}
\newline
\lmout{-"Levaquin" (active)}
}

\subsubsection*{Example 1: One-shot guided (incorrect)}

\exbox{[...] start her on Ativan 1 mg p.o. q. 8 hours and use Ativan 1 mg IV q. 4 hours p.r.n. for agitation. I will also start her on Inderal LA 60 mg p.o. q.d. for essential tremors. She does not want to take Celexa, and I will put her back on Lexapro 2 mg p.o. q.d. I will discontinue Esmolol.
\newline \newline
Create a bulleted list of which medications [...]
\newline \newline
-"Ativan" (discontinued) \newline
-"Inderal LA" (active) \newline
-"Celexa" (neither) \newline
-"Lexapro" (active) \newline
-"Esmolol" (active) \newline
\newline
[...] home dose of Kadian as this is her long-acting medication and DC the continuous Dilaudid given IV. 5. Urinary tract infection with Klebsiella and E. coli, both sensitive to Levaquin. Since this was diagnosed Foley has been DC'd. For now would continue Levaquin and recheck urinalysis.
\newline \newline
Create a bulleted list of which medications [...]
\newline \newline
-"\lmout{Kadian" (active)}
\newline
\lmout{-"Dilaudid" (discontinued)}
\newline
\lmout{-"Levaquin" (active)}
}

\subsubsection*{Example 1: One-shot guided (correct)}

\exbox{[...] start her on Ativan 1 mg p.o. q. 8 hours and use Ativan 1 mg IV q. 4 hours p.r.n. for agitation. I will also start her on Inderal LA 60 mg p.o. q.d. for essential tremors. She does not want to take Celexa, and I will put her back on Lexapro 2 mg p.o. q.d.
\newline \newline
Create a bulleted list of which medications [...]
\newline \newline
-"Ativan" (active) \newline
-"Inderal LA" (active) \newline
-"Lexapro" (active) \newline
-"Celexa" (neither)
\newline \newline
[...] home dose of Kadian as this is her long-acting medication and DC the continuous Dilaudid given IV. 5. Urinary tract infection with Klebsiella and E. coli, both sensitive to Levaquin. Since this was diagnosed Foley has been DC'd. For now would continue Levaquin and recheck urinalysis.
\newline \newline
Create a bulleted list of which medications [...]
\newline \newline
-"\lmout{Kadian" (discontinued)}
\newline
\lmout{-"Dilaudid" (discontinued)}
\newline
\lmout{-"Levaquin" (discontinued)}
}

\clearpage

\subsection{Medication Attribute Extraction}
\label{apx:medication_attribute_extraction}

\subsubsection*{Example 1: Token-level}

\exbox{<One-shot-example>
\newline \newline
Label the tokens: medication, dosage, route, frequency, duration, reason, or none. Ignore allergies.
\newline \newline
<One-shot-labels>
\newline \newline
Input: 8. Albuterol 2 puffs every 4-6 hours as needed. HOSPITAL COURSE: This is an 80-year-old female who was hospitalized about 2 months ago for chronic obstructive pulmonary disease exacerbation. At that time she was put on prednisone and antibiotics and seemed to get better. However, she was put on Augmentin ES and continued to have difficulty tasting food and felt that food tasted very salty. She had no appetite and she has continued to lose weight over the last 2 months.
\newline \newline
Label the tokens: medication, dosage, route, frequency, duration, reason, or none. Ignore allergies.
\newline \newline
-"\lmout{8": none} \newline
\lmout{-".": none} \newline
\lmout{-"Albuterol": medication} \newline
\lmout{-"2": dosage} \newline
\lmout{-"puffs": dosage} \newline
\lmout{-"every": frequency} \newline
\lmout{-"4-6": frequency} \newline
\lmout{-"hours": frequency} \newline
\lmout{-"as": none} \newline
\lmout{-"needed": none} \newline
\lmout{-".": none} \newline
\lmout{-"HOSPITAL": none} \newline
\lmout{-"COURSE": none} \newline
\lmout{-"This": none} \newline
\lmout{-"is": none} \newline
\lmout{-"an": none} \newline
\lmout{-"80-year-old": none} \newline
\lmout{-"female": none} \newline
\lmout{-"who": none} \newline
\lmout{-"was": none} \newline
\lmout{-"hospitalized": none} \newline
\lmout{-"about": duration} \newline
\lmout{-"2": duration} \newline
\lmout{-"months": duration} \newline
\lmout{-"ago": duration} \newline
\lmout{-"for": reason} \newline
\lmout{-"chronic": reason} \newline
\lmout{-"obstructive": reason} \newline
\lmout{-"pulmonary": reason} \newline
\lmout{-"disease": reason} \newline
\lmout{-"exacerbation": reason} \newline
\lmout{-".": none} \newline
\lmout{-"At": none} \newline
\lmout{-"that": none} \newline
\lmout{-"time": none} \newline
\lmout{-"she": none} \newline
\lmout{-"was": none} \newline
\lmout{-"put": none} \newline
\lmout{-"on": none} \newline
\lmout{-"prednisone": medication} \newline
[...]
}

\subsubsection*{Example 1: Phrase-level}

\exbox{<One-shot-example>
\newline \newline
Label the tokens: medication, dosage, route, frequency, duration, reason, or none. Chunk the same entity together. Ignore allergies.
\newline \newline
<One-shot-labels>
\newline \newline
Input: 8. Albuterol 2 puffs every 4-6 hours as needed. HOSPITAL COURSE: This is an 80-year-old female who was hospitalized about 2 months ago for chronic obstructive pulmonary disease exacerbation. At that time she was put on prednisone and antibiotics and seemed to get better. However, she was put on Augmentin ES and continued to have difficulty tasting food and felt that food tasted very salty. She had no appetite and she has continued to lose weight over the last 2 months.
\newline \newline
Label the tokens: medication, dosage, route, frequency, duration, reason, or none. Chunk the same entity together. Ignore allergies.
\newline \newline
-"\lmout{8": none} \newline
\lmout{-".": none} \newline
\lmout{-"Albuterol": medication} \newline
\lmout{-"2 puffs": dosage} \newline
\lmout{-"every 4-6 hours": frequency} \newline
\lmout{-"as needed": duration} \newline
\lmout{-".": none} \newline
\lmout{-"HOSPITAL COURSE": none} \newline
\lmout{-"This": none} \newline
\lmout{-"is": none} \newline
\lmout{-"an": none} \newline
\lmout{-"80-year-old": none} \newline
\lmout{-"female": none} \newline
\lmout{-"who": none} \newline
\lmout{-"was": none} \newline
\lmout{-"hospitalized": none} \newline
\lmout{-"about": none} \newline
\lmout{-"2 months": duration} \newline
\lmout{-"ago": none} \newline
\lmout{-"for": none} \newline
\lmout{-"chronic": none} \newline
\lmout{-"obstructive": none} \newline
\lmout{-"pulmonary": none} \newline
\lmout{-"disease": reason} \newline
\lmout{-"exacerbation": none} \newline
\lmout{-".": none} \newline
\lmout{-"At": none} \newline
\lmout{-"that": none} \newline
\lmout{-"time": none} \newline
\lmout{-"she": none} \newline
\lmout{-"was": none} \newline
\lmout{-"put": none} \newline
\lmout{-"on": none} \newline
\lmout{-"prednisone": medication} \newline
[...]
}

 \vspace{3cm}

\subsubsection*{Example 1: Relation Extraction}

\exbox{
<One-shot-example>
\newline \newline
Label medications, ignoring allergies. Include dosage, route, frequency, duration, reason, if available.
\newline \newline
<One-shot-labels>
\newline \newline
Input: 8. Albuterol 2 puffs every 4-6 hours as needed. HOSPITAL COURSE: This is an 80-year-old female who was hospitalized about 2 months ago for chronic obstructive pulmonary disease exacerbation. At that time she was put on prednisone and antibiotics and seemed to get better. However, she was put on Augmentin ES and continued to have difficulty tasting food and felt that food tasted very salty. She had no appetite and she has continued to lose weight over the last 2 months.
\newline \newline
Label medications, ignoring allergies. Include dosage, route, frequency, duration, reason, if available.
\newline \newline
-"\lmout{medication: "Albuterol", dosage: "2 puffs", frequency: "every 4-6 hours", duration: "as needed"} \newline
\lmout{-medication: "prednisone", duration: "2 months"} \newline
\lmout{-medication: "antibiotics", duration: "2 months"} \newline
\lmout{-medication: "Augmentin ES", duration: "2 months"} \newline
}

%% file: 11_appendix_annotation.tex
\clearpage 
\begin{figure*}[ht!]
    \centering
    \includegraphics[width=0.9\textwidth]{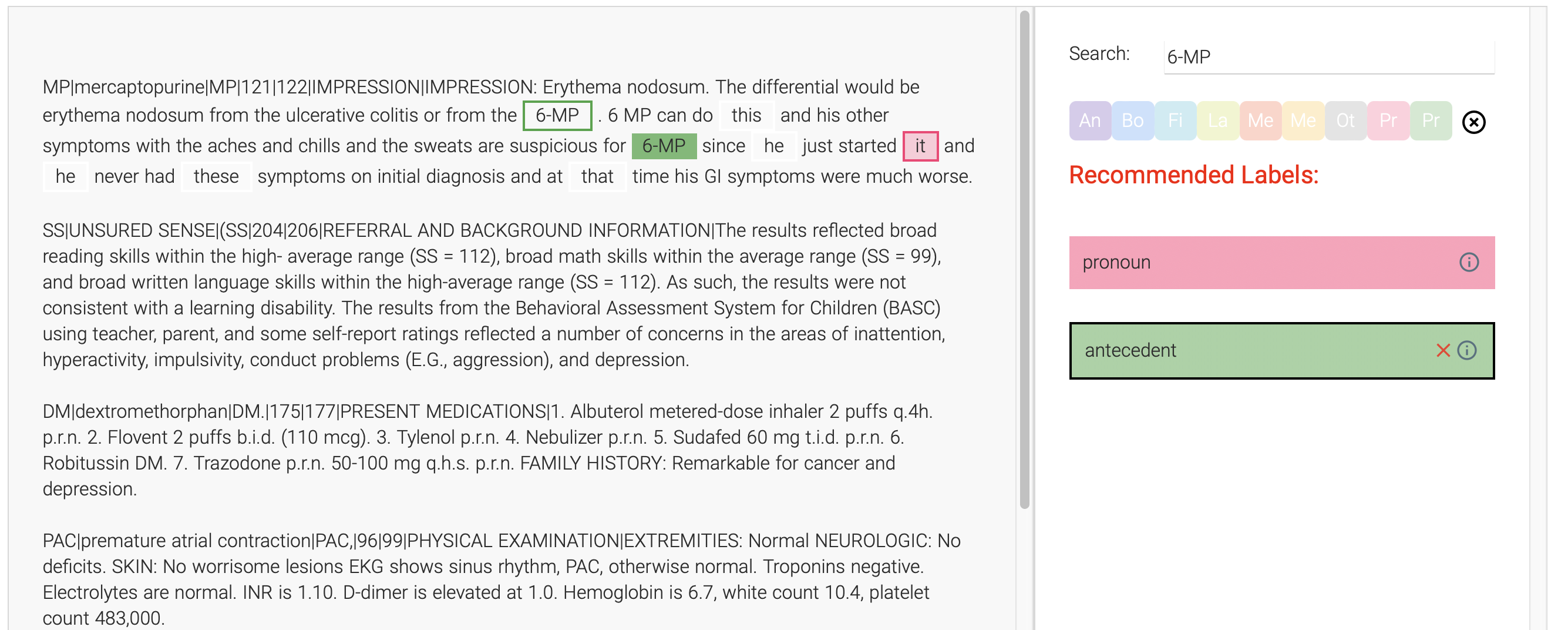}
    \caption{Platform used for annotation of the three new labeled datasets, shown for the coreference resolution annotation task. }
    \label{fig:annotation}
\end{figure*}

\section{Annotation Process}
\label{apx:annotation_process}
A screenshot of the annotation process can be seen in Figure 2.
\subsection{Biomedical Evidence Extraction}

For arm identification 20 out of 187 abstracts from the test set of \citet{nye2018corpus} were randomly selected. 
Both annotators separately identified the study arms in each abstract.
They included all characteristics of a arm that were necessary for differentiation \cite{ferracane2016leveraging}.
For example, we would not require the route of administration for a drug (e.g. “oral" in \textit{oral X}) unless another arm contained the the same drug in a different formal (e.g. \textit{X nasal spray}).
There was full consensus between annotators for the identified numbers of arms.
A single abstract was replaced due to its ambiguity.

\subsection{Coreference Resolution}
\label{apx:coreference-resolution-dataset-details}
Annotators labeled 105 snippets from the CASI dataset with pronouns and their corresponding noun phrase antecedent \cite{moon2014sense}. The antecedent was annotated as the entire noun phrase (barring any dependent clauses); in cases where two antecedents were available, both were labeled. For the purposes of evaluation, we chose the antecedent with the highest overlap to each model's output.
To ensure nontrivial examples, the annotators excluded all examples of personal pronouns (e.g. ``he'', ``she'') if another person (and possible antecedent) had not yet been mentioned in the snippet.

\subsection{Medication Status Extraction}
\label{apx:medication-status-extraction-dataset-details}
We wanted to create a dataset of challenging examples containing a changeover in treatment.
From a sample, only $\sim$5\% of CASI snippets contained such examples.
To increase the density of these  examples, speeding up annotation, clinical notes were filtered with the following search terms: \textit{discont}, \textit{adverse}, \textit{side effect}, \textit{switch}, and \textit{dosage}, leading to 1445 snippets.
We excluded snippets that were purely medication lists, requiring at least some narrative part to be present.
For 105 randomly selected snippets, the annotators first extracted all medications.
Guidelines excluded medication categories (e.g. ``ACE-inhibitor'') if they referred to more specific drug names mentioned elsewhere (even if partially cut off in the snippet).
For instance, only the antibiotic Levaquin was labeled in: ``It is probably reasonable to treat with antibiotics [...]. I would agree with Levaquin alone [...]''. 
Guidelines also excluded electrolytes and intravenous fluids as well as route and dosage information.
In a second step, medication were assigned to one of three categories: \textit{active}, \textit{discontinued}, and \textit{neither}. 
Discontinued medications also contain medications that are temporarily on hold.
The category \textit{neither} was assigned to all remaining medications (e.g. allergies, potential medications).

\subsection{Medication Attribute Extraction}
\label{apx:med-attribute-annotation}
For medication attribute extraction, we also labeled 105 examples from CASI \cite{moon2014sense}. 
Annotation guideline were adopted from the 2009 i2b2 medication extraction challenge \cite{uzuner2010extracting} with slight modifications.
We allowed medication attributes to have multiple spans.
Also, we grouped together different names of the same drug (e.g. ``Tylenol'' and ``Tylenol PM'') for the purpose of relation extraction.
After annotation of the data, we create three versions of the dataset: token-level, phrase-level, and relation-level.
For the first, we split all word in the example and assigned them their respective label or \emph{none} if they were not part of a label (see token-level example in \ref{apx:medication_attribute_extraction}.
For phrase-level, we kept consecutive words with the same label grouped together as phrases (see phrase-level example in \ref{apx:medication_attribute_extraction}.
The relation level just contained the extracted medication and their attributes (see relation extraction example in \ref{apx:medication_attribute_extraction}.
We note that medication lists were downsampled in the creation of the dataset, since the 2009 i2b2 challenge had found performance on narrative text was far lower than on medication lists.

%% file: 12_annotation_exp_details.tex
\clearpage 
\clearpage
\section{Additional Experimental details}
\label{apx:experimental_details}
Across all datasets, similar to \citet{honovich2022instruction}, we assume the latest generation of models on the OpenAI API is the set of InstructGPT models from \citet{ouyang2022training}. 
\subsection{Clinical Sense Disambiguation}
\label{sec:casi-not-in-training-set}

\textbf{How do we know CASI is not in the LLM training set?}
Since the CASI dataset is publicly accessible from the Internet and on Github, one potential pitfall is that the dataset may have been in the language models' training data. 
While this is also true of other common NLP benchmarks, we attempted to confirm results were not merely an artifact. 
To do so, we annotated 50 distinct acronyms that occurred in sentences in the CASI dataset that were \textit{not} included in the original annotations. 
While this set of acronyms is easier (e.g., they many only have a single clinical expansion), this allows us to check that GPT-3 is not simply pattern matching to potential past training data. 
In the set of 50, we find \textit{GPT-3 edit} correctly expanded 47 (94\%).
In 2 of these cases, the acronym was in fact a typo (SMIV instead of SIMV, AVG instead of ABG), and the correct expansion was given regardless. 
Of the 3 that were incorrect, one was in fact incorrect, one was of unspecified meaning to the annotator, and one had 2/3 of the words correct in the expansion.
\textbf{Resolver Details}

\textbf{Weak Supervision}
For weak supervision, we only consider the 97\% of the dataset where the overlap with an answer choice was at least 5 characters as candidates for pseudolabels.
Following prior work \cite{lang2022co, lang2022training}, we additionally used a technique called the \emph{cut statistic} to select a high-quality subset of the weakly labeled data to reduce the noise in the training process.
We selected a subset of size 75\% to decrease noise while still choosing a large enough set to ensure all acronyms were seen during training.
We fine-tuned a PubMedBERT \cite{gu2021domain} model, a BERT variant that was pretrained on biomedical abstracts and full-text articles from PubMed, using learning rate 1e-5, weight decay 0.01, the AdamW optimizer \citep{loshchilov2018decoupled}, and batch size 4, using the \texttt{BERTForMultipleChoice} functionality in HuggingFace Transformers \cite{wolf2019huggingface}.

\subsection{Biomedical Evidence Extraction}
\label{apx:biomedical-evidence-extraction}
\paragraph{Baseline Training Details}
We trained for 10,000 steps using the AdamW optimizer, learning rate 2e-5, batch size 32, and weight decay 1e-6, inheriting these hyperparameters from \citet{zhang2021wrench}. 
These were the best-performing hyperparameters across the set reported in \citet[][Table 10, ``BERT-CRF'']{zhang2021wrench}.

\paragraph{Resolver Details}
To evaluate on the original token-level labels we tokenize the GPT-3 output and remove bullet points, numbers, stop words, and the words ``treatment'', ``control'', and ``group'' which GPT-3 often appended for clarification (e.g. ``- Placebo (Control group)'').
Then, any token in the input that is found in the remaining GPT-3 output is labeled with a 1, and others with a 0. 
Since our procedure may have interrupted valid spans, we fill in any 0's between 1's as well as acronyms within parentheses. 
These steps transform the LLM output strings $l_i$ to a binary labeling of the full input.

\paragraph{Example of Token-level Error Modes}

As an example describing token-level error modes of GPT-3, consider the output, the resolved output, and the gold label for a study with two arms below. 

\noindent  \ul{GPT-3 output}\\
\textit{- Inhaled fluticasone\\
- Placebo} \\

\noindent \ul{Resolved GPT-3 output}:

\noindent \textit{\textcolor{brown}{Inhaled fluticasone} reduces [...] double-blind, placebo-controlled study [...] \textcolor{brown}{inhaled fluticasone} [...] or \textcolor{brown}{placebo.} Large-scale [...] of inhaled steroid therapy on [...]} \\

\noindent \ul{Gold-label (token-level)}:

\noindent \textit{Inhaled \textcolor{brown}{fluticasone} reduces [...] double-blind, \textcolor{brown}{placebo-controlled} study [...] inhaled \textcolor{brown}{fluticasone} [...] or \textcolor{brown}{placebo.} Large-scale [...] of inhaled \textcolor{brown}{steroid} therapy on [...]}\\

\noindent GPT-3 correctly identifies both study arms. 
However, the resolved output, which simply labels the token sequence of the identified arms in the original input, disagrees with the gold labels for several tokens. For example, the output includes the route, ``inhaled'', which isn't kept in the annotation schema, dinging precision. Further, the output excludes ``placebo-controlled'' (given ``placebo'' is included), dinging recall.  Therefore, despite qualitatively capturing the arms of this trial, there was a middling F1-score of 0.70 for this example. This serves to underline why token-level metrics can be misleading as to true performance towards the underlying goal.

\paragraph{Oracle Details}
We assumed oracle splitting and oracle coreference resolution in order to distill the token-level labels to a list for the PubMedBERT baselines.
As an example of oracle splitting, PubMedBERT assigned a $1$ to the span \textit{``40, 120, and 240 s exposure to 3,000, 700, and 500mg l$^1$ clove solution;''} this span in fact contains three different arms, and we assume it can be perfectly split, since the required information is theoretically present in the identified span. 
As an example of oracle coreference resolution, consider this example with two arms: \textit{capecitabine and oxaliplatin plus radiotherapy (Cap-Oxa-CRT)} and \textit{concurrent capecitabine and radiotherapy (Cap-CRT)}. 
The spans recognized by PubMedBERT include {``adjuvant concurrent chemotherapy''}, {``capecitabine-based concurrent chemotherapy''}, {``postoperative CRT of capecitabine with or without oxaliplatin''}, {``concurrent capecitabine and radiotherapy (Cap-CRT)''} and {``capecitabine and oxaliplatin plus radiotherapy (Cap-Oxa-CRT).''}
To be generous to the baseline, we assumed those 5 spans \textit{could} possibly be reduced to the two arms with oracle coreference resolution. No oracle splitting or coreference resolution was conducted for \method.

\paragraph{Analysis of Error Modes for Arm Identification}
\method{} successfully identified the correct number and content of the arms in 17 of the 20 examples. 
The three examples it missed were also missed by PubMedBERT. 
In one case with two arms, both methods included a procedure as a separate third arm; in reality, the procedure occurred for both arms and was not the intervention itself.
In the second case, the prompt output did not elaborate on the treatment group sufficiently, and in the final case, it fully misparsed. 
Assuming the oracle splitting and coreference, PubMedBERT would still have issues with 10 further examples: two again included a common procedure as a third arm, four were missing control arms, one was missing a treatment arm, two arms required further domain knowledge to consolidate (e.g., that Ramipril is an ACE inhibitory therapy), and another required properly consolidating a therapy with no overlapping tokens.

\subsection{Coreference Resolution}
\paragraph{Baseline Details}
We benchmark using a transformer-based model trained jointly on three large coreference datasets \cite{toshniwal2021generalization} that can be found on the HuggingFace model hub (\texttt{shtoshni/longformer\_coreference\_joint}).
\paragraph{Resolvers}
The resolver for the 0-shot unguided prompt was 50 LOC, or 973 tokens in the Codex tokenizer. In contrast, the 1-shot guided prompt required only stripping a final quotation mark, period, or space, which required 20 tokens per the Codex tokenizer.
\subsection{Medication + Status Extraction}
\label{apx:medication-status-extraction-resolver}
\textbf{Resolver details} 
For an unguided prompt, to map the GPT-3 output string to a list of medication strings, the first step is to break the output string up into substrings by parsing the ``bulleted list'' output by GPT-3, which we do with regular expressions.
The output strings for this prompt followed several different formats, making this step slightly more involved than in previous cases.
The two basic formats were a newline-separated list and a comma-separated list of medication names.
The modifiers were also expressed in different ways: some outputs were \textit{\{Medication\}: \{Status\}}, while others were \textit{\{Medication\} (\{Status\})}.
A few examples instead grouped the medications by status, so the output was \textit{Active: \{medication1\}, \{medication2\}, Discontinued: \{medication3\}}.
Examples of these outputs can be found in Appendix \ref{apx:medication_status_extraction}.
Despite this variation, we output a list by simply replacing newlines with commas to reduce to the comma-separated case, and then applying two regular expressions to extract the medication names and modifiers from the list.

The previous steps turn the LLM output strings into lists of strings.
The next step in the resolver is to \emph{denoise} the individual strings in each list by first stripping dosage and route information (e.g., ``10 mg'' or ``patch'') and then performing input-consistency checking by removing tokens that do not appear in the input.
Finally, strings that, after the prior denoising steps, only consist of stop words or primarily consist of punctuation and whitespace, are removed from the prediction lists.
This required 32 lines of code, and 946 tokens in a byte-pair encoding. In contrast, with a 1-shot prompt, output could be simply split on the bullets, and the status extracted from parentheses, requiring 8 lines of code and 165 tokens in a byte-pair encoding. 

\paragraph{Medication Extraction Baseline}
For normalization, all entities were linked to the UMLS via the default string overlap functionality of ScispaCy \cite{bodenreider2004unified}.
We filtered the resulting UMLS concepts by their semantic types and only kept concepts of the types \textit{Antibiotic}, \textit{Clinical Drug}, \textit{Pharmacologic Substance}, and \textit{Vitamin}.
Finally, the baseline predictions are run through the same denoising steps as the GPT-3 predictions to ensure a fair comparison.

\paragraph{Status Classification: T-Few}
We use T-Few \citep{liu2022few} for medication status classification using 20 additional annotated examples as the few-shot training set.
We used a single prompt:
\vspace{1mm}
\exbox{In the clinical note below, what is the status of the medication Albuterol?
\newline \newline
Albuterol 2 puffs every 4-6 hours as needed. HOSPITAL COURSE: This is an 80-year-old female who was hospitalized about 2 months ago for chronic obstructive pulmonary disease exacerbation. At that time she was put on prednisone and antibiotics and seemed to get better. However, she was put on Augmentin ES and continued to have difficulty tasting food and felt that food tasted very salty. She had no appetite and she has continued to lose weight over the last 2 months.
}

For the answer choices, we used \texttt{\small Discontinued}, \texttt{\small Active}, and \texttt{\small Neither}. 
We did not use IA$^{(3)}$ pretraining, but otherwise directly followed the T-Few recipe (i.e., we used the default values for all hyperparameters including batch size, learning rate, number of steps, length normalization, etc.).
We used the T0-11B model.
\subsection{Medication + Relation Extraction}
\label{apx:medication-baselines}
\paragraph{Resolver}
The resolver for the first two tasks iterates over the lines in the GPT-3 output and grabs both the text span and the label; the text span is mapped to tokenized space, and all labels not in the label space (e.g. ``Instructions'') are mapped to \textit{None}. For phrase-level labeling, a single additional step is conducted to map the labels to BIO format. For the relation extraction task, the resolver additionally assumes all entities mentioned in a line correspond to the medication on that line.

\paragraph{Sequence Tagging baseline}
We model extraction and labeling of medication + modifier (dosage, frequency, route, duration, reason) as a sequence tagging task.
We use the \texttt{B/I/O} encoding for the label space, adding tags to the \texttt{B} and \texttt{I} labels indicating the type of entity.
For training data, we split the 10 notes from the 2009 i2b2 challenge into shorter contexts using an off-the-shelf sentence segmenter, and merged split contexts of less than 30 tokens into the previous context.
This results in 176 training contexts for the PubMedBERT + CRF model.
As with Biomedical Evidence Extraction, we search for hyperparameters over the search space reported in \citet[][Table 10, ``BERT-CRF'']{zhang2021wrench}.
The final model is chosen based on validation F1 score on a randomly selected validation set of 10\% of the training data (i.e., 18 contexts).

\paragraph{Relation Extraction Baseline}
We use the model from \citet{shi2019simple} for relation extraction on top of PubMedBERT.
For training data, we again use the 2009 i2b2 challenge set, but since the goal is to associate modifiers with individual medications, we split up the 10 long notes into rolling chunks around each medication mention.
For each ground-truth medication entity, we create a context including the 30 tokens before and after that entity.
We extended these windows to be on an \texttt{O} label so that entities are not split across contexts.
We use a binary label space, since each modifier type (dosage, route, etc.) determines the relation type: the relevant task is to classify whether each pair of (medication, modifier) entities in a span is associated.
We create one positive sample for each truly related (medication, modifier) pair.
For each context, we add a negative sample for each (medication, modifier) pair that is not related.
This results in 1416 examples (many of which have largely overlapping context, but a different pair of entities) for training the relation extraction model.

%% file: 13_appendix_cost.tex
\clearpage

\onecolumn
\begin{table}[tb!]
\centering
\small
\resizebox{0.8\textwidth}{!}{\begin{tabular}{|P{1.8cm}|P{2.2cm}|P{2cm}|P{2cm}|P{2cm}|P{2cm}|}
\hline
\textbf{Task} &
  \textbf{Cost/Token} &  \textbf{Tokens/Example} & \textbf{\# of Examples} & \textbf{Experimental Settings} & \textbf{Estimated Cost} \\ \hline\hline
Clinical sense \mbox{disambiguation} &
  \$0 (free in \textit{edit} beta mode)  &
  100 
  & 105 &
  1
  & \$0\\ \hline
Biomedical evidence extraction &
  \$0.00006 &
 500 & 187 & 1 & \$6 \\ \hline
Coreference resolution &
   \$0.00006 &
 300 & 105 & 11 & \$21 \\ \hline
Medication status extraction &
     \$0.00006 &
 300 & 105 & 16 & \$30\\ \hline
Medication attribute extraction &
    \$0.00006 &
 600 & 105 & 3 & \$12 \\ \hline
\end{tabular}}
\caption{Estimate of cost of running the experiments included in this work}
\label{tab:tasks_cost}
\end{table}

\section{Experimental Cost}
\label{apx:experimental_cost}
At time of experimentation the cost of experiments included in this work were under \$100.
A breakdown of the upper bound of API costs can be found in the table below and is based on OpenAI API pricing in spring 2022. All estimates of tokens/example are rough upper bounds; some experimental settings were cheaper.
